\documentclass[manuscript]{acmart}

\AtBeginDocument{%
	\providecommand\BibTeX{{%
			\normalfont B\kern-0.5em{\scshape i\kern-0.25em b}\kern-0.8em\TeX}}}

\setcopyright{acmcopyright}

\copyrightyear{2021} 
\acmYear{2021} 
\setcopyright{acmcopyright}\acmConference[IUI '21]{26th International Conference on Intelligent User Interfaces}{April 14--17, 2021}{College Station, TX, USA}
\acmBooktitle{26th International Conference on Intelligent User Interfaces (IUI '21), April 14--17, 2021, College Station, TX, USA}
\acmPrice{15.00}
\acmDOI{10.1145/3397481.3450655}
\acmISBN{978-1-4503-8017-1/21/04}

\usepackage{amsmath}
\usepackage{graphicx}
\usepackage[T1]{fontenc}   
\usepackage{ccicons}          
\usepackage{paralist}
\usepackage{hyperref}
\usepackage[nolist]{acronym}
\usepackage{color}

\def\hb{\hbox to 10.7 cm{}}

\newcommand{\quotes}[1]{``#1''}

\begin{document}
		
	\title{From Philosophy to Interfaces: an Explanatory Method and a Tool Inspired by Achinstein’s Theory of Explanation}
	
	\author{Francesco Sovrano}
	\email{francesco.sovrano2@unibo.it}
	\orcid{0000-0002-6285-1041}
	\author{Fabio Vitali}
	\email{fabio.vitali@unibo.it}
	\orcid{0000-0002-7562-5203}
	
	\renewcommand{\shortauthors}{Sovrano, et al.}
	\renewcommand{\shorttitle}{From Philosophy to Interfaces}
	
	\begin{abstract}
		We propose a new method for explanations in \ac{AI} and a tool to test its expressive power within a user interface. In order to bridge the gap between philosophy and human-computer interfaces, we show a new approach for the generation of interactive explanations based on a sophisticated pipeline of \ac{AI} algorithms for structuring natural language documents into knowledge graphs, answering questions effectively and satisfactorily. 
		Among the mainstream philosophical theories of explanation we identified one that in our view is more easily applicable as a practical model for user-centric tools: \citeauthor{achinstein1983nature}'s \textit{Theory of Explanation}.
		With this work we aim to prove that the theory proposed by \citeauthor{achinstein1983nature} can be actually adapted for being implemented into a concrete software application, as an interactive process answering questions.
		To this end we found a way to handle the generic (\textit{archetypal}) questions that implicitly characterise an explanatory processes as preliminary overviews rather than as answers to explicit questions, as commonly understood. To show the expressive power of this approach we designed and implemented a pipeline of \ac{AI} algorithms for the generation of interactive explanations under the form of overviews, focusing on this aspect of explanations rather than on existing interfaces and presentation logic layers for question answering.
		Accordingly, through the identification of a minimal set of archetypal questions it is possible to create a generator of explanatory overviews that is generic enough to significantly ease the acquisition of knowledge by humans, regardless of the specificities of the users outside of a minimum set of very broad requirements (e.g. people able to read and understand English and capable of performing basic common-sense reasoning).
		We tested our hypothesis on a well-known XAI-powered credit approval system by IBM, comparing CEM, a static explanatory tool for \textit{post-hoc} explanations, with an extension we developed adding interactive explanations based on our model.
		The results of the user study, involving more than 100 participants, showed that our proposed solution produced a \textit{statistically relevant} improvement on effectiveness (U=931.0, p=0.036) over the baseline, thus giving evidence in favour of our theory.
	\end{abstract}

	
		
	\maketitle
	
	\section{Introduction} \label{sec:introduction}
	The complexity of modern software and the increasing discomfort of humans towards the correctness and fairness of the output of such complex systems has caused the birth and growth of a new discipline to reduce the distance between individuals, society, and machines: \ac{XAI}.
	Governments have also started to act towards the establishment of ground rules of behaviour from complex systems, for instance through the enactment of the European \acf{GDPR} (2016\footnote{Regulation (EU) 2016/679.}), which identifies \emph{fairness}, \emph{lawfulness}, and in particular \emph{transparency} as basic principles for every data processing tools handling personal data; even creating a new \emph{Right to Explanation} for individuals whose legal status is affected by a solely-automated decision. 
	The \acf{AI-HLEG} was established in 2018 with the explicit purpose of applying the principles of the \ac{GDPR} specifically to \ac{AI} software, and produced a list of fundamental ethical principles for \emph{Trustworthy \ac{AI}} tools that include \emph{fairness} and \emph{explicability}. \ac{XAI}, therefore, is understood as a way to answer these fundamental principles in the field of \ac{AI}. 
	Among the characteristics that the \ac{AI-HLEG} has identified, there is the concept of \emph{user-centrality}. In order to address such concept, explanations have to cater to the individuals in terms of their background, context, needs and purposes: a single explanation output (i.e., a \emph{One-Size-Fits-All} explanation) regardless of all these backgrounds and needs would become massive and unwieldy as soon as the data and processes to be explained become averagely complex, and the readers approaching such explanations would be immediately overwhelmed by the sheer size of the data. Consequently, One-Size-Fits-All approaches to explanations cannot be considered appropriate to this end as soon as the complexity of these systems surpasses a fairly trivial threshold, and a more sophisticated approach needs to be identified. 
	Despite these considerations, it appears that \ac{XAI} techniques are mainly focused on pursuing One-Size-Fits-All explanations, justified by convenient definitions framing an explanation as the product of an act of making things explainable rather than a pragmatic (user-centric) act of explaining based on explainability.
	
	In this paper, we propose a new approach to explanations in Artificial Intelligence. 
	We are interested in modelling an explanatory process for producing user-centric explanatory software and quantifying the difference it bears in terms of effectiveness with respect to non-pragmatic approaches. 
	More precisely, we want to understand how to structure information in order to facilitate the production of explanations of complex decision-making processes (such as explainable artificial intelligence processes).
	We acknowledge that we are not the first to try to model an explanatory process. In literature there were various efforts in this direction and a long history of debates and philosophical traditions, often rooted in Aristotle's works and those of other philosophers. Among the many models proposed over the last few centuries some are now considered fallacious, albeit historically useful (e.g. \citeauthor{hempel1965aspects}'s one \cite{hempel1965aspects}), but we have identified the one that, we believe, is mostly convertible into a practical model for user-centric explanatory software: \citeauthor{achinstein1983nature}'s \textit{illocutionary} theory of explanations, as answers to questions.
	Despite its criticalities, \citeauthor{achinstein1983nature}'s theory seemed to us the most suitable among all, for our purposes, for allowing the assessment of the quality of an explanation/answer on the base of its \textit{pragmatic} relevance to a question. A task that may seem too onerous and subjective, nonetheless recent developments in modern artificial intelligence have shown there might exist tools \cite{yang2019multilingual,roy2020lareqa} to objectively estimate the pertinence of an answer thus allowing for the automation of a question answering process.
	
	Though, modelling an explanatory process as a standard \ac{QA} process gave us the first impression of being a little bit unrealistic. Think of the following example of the \quotes{university lectures}: students (the explainees) follow the lessons to acquire (initially obscure) information provided by the professor (the explainer). A lesson can normally include the intervention of students in the form of observations and/or questions, but these interventions are, in practice, always after an initial phase of information acquisition. In other terms, the initial overview given by the professor may not be the answer to a preliminary question (especially if the students know absolutely nothing about what the professor is supposed to say). Regardless this apparent lack of a question, we might all agree that the professor could actually explain something good to the students.
	At this point it would seem that \citeauthor{achinstein1983nature}'s theory, being heavily based on question-answering, fails to capture the need for preliminary overviews during an explanatory process, as in the \quotes{university lectures} example. 
	Despite this first impression, we think that overviews can be generated as answers as well, therefore partially confirming \citeauthor{achinstein1983nature}'s original theory. 
	In fact, for the generation of an overview it is necessary to select and group information appropriately, so as to facilitate the production of different explanatory paths for different users. As analogy, we might see the space of all explanations (or explanatory space \cite{sovrano2020modelling}) as a sort of manifold where every single overview is not user-centred locally, but globally (as an element of a sequence of information, chosen by a user).
	What we theorize is that it is possible to generate such overviews through the identification of a set of archetypal questions (e.g. Why? What for? How? When? etc..) that allow contextualised information to be grouped and filtered according to its relative pertinence with respect to the archetypes. 
	If our hypothesis is correct, through the identification of a minimal set of archetypal questions, it is possible to obtain a generator of explanatory overviews generic enough to be able to significantly ease the acquisition of knowledge, regardless of the specific user but depending instead on a category of selected users (e.g. people able to read and understand English, hence capable of performing common-sense reasoning), thus resulting in a user-centred explanatory tool that is more effective than its non-pragmatic counterpart based on the same informative contents.

	In order to test our hypothesis, we had to:
	\begin{inparaenum}[i)]
		\item invent a new pipeline of \ac{AI} algorithms for the generation of overviews,
		\item design a simple user interface for presenting the overviews,
		\item and run a user study on it.
	\end{inparaenum}
	The pipeline is meant to organize the information contained in non-structured documents written in natural language (web pages, pdf, etc..), allowing efficient information clustering, according to a set of archetypal questions, aiming to build a sufficiently rich and effectively explorable \acf{ES} for the automated generation of user-centred explanations. 
	
	This paper is structured as follows. In Section \ref{sec:background} we provide a brief introduction to the contemporary philosophical developments in the theory of explanation, focusing on \citeauthor{achinstein1983nature}'s one.
	In Section \ref{sec:proposed_solution} we describe our proposed solution going through the details of the pipeline, thus comparing our work with existing literature in Section \ref{sec:related_work}.
	In Section \ref{sec:proof_of_concept} we present a proof of concept evaluation in the form of an user study on a XAI-powered credit approval system augmented with a simple extension for the navigation of the \ac{ES}. In Section \ref{sec:results} we show and discuss the obtained results, drawing the conclusions in Section \ref{sec:conclusions}.
	
	\section{Background: Contemporary Developments in the Theory of Explanation} \label{sec:background}
	In philosophy the terms \quotes{truth} and \quotes{explanation} have divergent interpretations \cite{mayes2005theories}. 
	On one hand, according to the realist interpretation \quotes{the truth and explanatory power of a theory are matters of the correspondence of language with an external reality}, so it is more about depicting reality with words and intents. 
	On the other hand, the epistemic interpretation defines explanation as a mere re-ordering of phenomena and experience to a greater degree, focusing on its power to explain observable phenomena rather than its literal truth.
	The failure to distinguish these senses of \quotes{explanation} can and does foster disagreements that are purely semantic in nature.
	
	According to \citeauthor{mayes2005theories} \cite{mayes2005theories}, explanation in philosophy has been conceived within the following five traditions: 
	\begin{itemize}
		\item \textbf{Causal Realism} \cite{salmon1984scientific}: explanation as the articulation of the fundamental causal mechanisms of a phenomenon.
		\item \textbf{Constructive Empiricism} \cite{van1980scientific}: epistemic theory of explanation that draws on the logic of why-questions and on a Bayesian interpretation of probability.
		\item \textbf{Ordinary Language Philosophy} \cite{achinstein1983nature}: the act of explanation as the attempt to produce understanding in another by answering a certain kind of question in a certain kind of way.
		\item \textbf{Cognitive Science} \cite{holland1989induction}: explaining as a process of belief revision, etc..
		\item \textbf{Naturalism and Scientific Realism} \cite{sellars1963philosophy}: rejects any kind of explanation of natural phenomena that makes essential reference to unnatural phenomena. Explanation is not something that occurs on the basis of pre-confirmed truths. Rather, successful explanation is actually part of the process of confirmation itself. 
	\end{itemize}

	\subsection{Achinstein's Theory of Explanations}
	In \citeyear{achinstein1983nature}, \citeauthor{achinstein1983nature} \cite{achinstein1983nature} was one of the first to analyse the whole process of generating explanations, introducing his philosophical model of a \textit{pragmatic} explanatory process. According to his model, explaining is an illocutionary act coming from a clear intention of producing new understandings in an explainee by providing a correct content-giving answer to an open question. 
	Therefore, according to this view, answering by \quotes{filling the blank} of a pre-defined template answer (as most of One-Size-Fits-All approaches do) prevents the act of answering from being explanatory, by lacking illocution. These conclusions are quite clear and explicit in \citeauthor{achinstein1983nature}'s last works \cite{achinstein2010evidence}, consolidated after a few decades of public debates.
	More precisely, according to \citeauthor{achinstein2010evidence}'s theory, an explanation can be summarized as a pragmatically correct content-giving answer to one or more questions of various kinds, not necessarily linked to causality.
	As consequence we can see a deliberate absence of a taxonomy of questions to refer to (helpful to categorize and better understand the nature of human explanations).
	This results in a refusal to define a quantitative way to measure how pertinent an answer is to a question, justified by the important assertion that explanations have a pragmatic character, so that what exactly has to be done to make something understandable to someone may (in the most generic case) depend on the interests and background knowledge of the person seeking understanding \cite{douven2012peter}.
	In this sense, the strong connection of \citeauthor{achinstein2010evidence}'s theory to natural language and natural users is quite evident, for example in the \citeauthor{achinstein2010evidence}ian concept of \textit{elliptical understandings} as \quotes{understandings of what significance or importance X has (in the present context)} \cite{achinstein2010evidence} or in the concept of \textit{u-restrictions} where an utterance/explanation can be said to express a proposition if and only if it can appear in (many) contexts reasonably known by the explainee.
	But, despite this, \citeauthor{achinstein2010evidence} does not reject at all the utility of formalisms, hence suggesting the importance of following \textit{instructions} (protocols, rules, algorithms) for correctly explaining some specific things within specific contexts.
	
	\section{Proposed Solution} \label{sec:proposed_solution}
	\citeyear{achinstein1983nature}'s illocutionary theory of explanation is pragmatic, user-centric, but it seems very hard to be concretised into a software.
	The point is: what is \textit{illocution}, precisely? Although also the dictionary gives its own definition, this seems to be too abstract to be implementable into a real user-centric explanatory software, also considering that in \citeauthor{achinstein1983nature}'s theory there is a deliberate absence of a taxonomy of questions, that might be needed for a more objective evaluation of the quality of an explanation.
	Despite its well-known criticalities, \citeauthor{achinstein1983nature}'s model seemed to us the most suitable among all for building a user-centric explanatory software, by suggesting that explaining is akin to \acf{QA}. 
	On this aspect, one of the main technological limitations of state-of-the-art automated \ac{QA} \cite{yang2019multilingual,roy2020lareqa} is that it tends to lose its effectiveness when the questions are too broad. This results in several issues when the user has no knowledge of the domain, thus forcing him/her to resort to generic questions in order to acquire enough information to be able to approach more specific questions. Seemingly, this technological issue may have its own counterpart in \citeauthor{achinstein1983nature}'s theory as well, as shown in the \quotes{university lectures} example presented in Section \ref{sec:introduction}. 
	In fact, it would seem that \citeauthor{achinstein1983nature}'s theory, being so heavily based on question-answering, fails to capture the need for preliminary overviews during an explanatory process. 
	Despite this first impression, we think that overviews can be generated as answers as well, therefore partially confirming \citeauthor{achinstein1983nature}'s original theory. 
	%
	Indeed, we believe that illocution in explanations is equivalent to the act of \textit{pertinently answering} \textit{archetypal questions} (e.g. Why? What for? How? When? etc..), and that is different from the \citeauthor{achinstein2010evidence}'s concept of \textit{instructions} but in some way akin to the \citeauthor{achinstein2010evidence}ian concept of \textit{elliptical understandings} briefly introduced in Section \ref{sec:background}. In fact, explaining is not just answering a given question in a punctual way (that would simply be answering) but it is also answering all the other implicit (archetypal) questions defined by: the explainee's background knowledge, the objectives of the explanatory process and the given context. It is, in some sense, attempting to anticipate the (conceivably mostly unknown) explainee's needs for an explanation by providing, as an archetypal answer, possibly expandable summaries of (more punctual) pertinent information. 
	In other terms, the more archetypal answers about the explanandum's aspects are covered by the act of explaining, the more likely the resulting explanation is going to meet the explainee's objectives, the better is the explanatory process.
	The archetypal questions prevent by design any \quotes{filling the blank} answer, thus meeting the tricky but reasonable assumption of illocution given by \citeauthor{achinstein1983nature} for his pragmatic theory of explanations. What is of most interest is that, perhaps, the assumption of explaining as \textit{pertinently answering} also to \textit{archetypal questions} is simpler, stronger and more precise than the assumption of illocution, removing from the \quotes{equation} (therefore simplifying it) the need for a model of human intention.
	Nonetheless, one main technical issue with this theory of explanations is with respect to user-centrality, requiring an explainer (e.g. an explanatory system) to actually anticipate the explainee's (non-explicit) needs; how could one estimate pertinence without exactly knowing with respect to what? 
	In a context in which only very minimal assumptions can be made on the explainee, we propose one more strong difference with \citeauthor{achinstein2010evidence}'s model. In this context, we define \textit{pertinently answering} (to archetypal questions) as the process of giving (archetypal) answers that are likely to be pertinent for a given (archetypal) question. The likelihood can be quantitatively estimated on strong-enough statistical evidence collected from large corpora and built in language models. This statistical definition of pertinence is also compatible with the definition of \textit{u-restrictions} given by \citeauthor{achinstein2010evidence} and it does not preclude a pragmatic (user-centred) explanatory process, as we are going to show, that is locally non-pragmatic but globally pragmatic.
	In fact, for the generation of an overview it is necessary to select and group information appropriately, so as to facilitate the production of different explanatory paths for different users. As analogy, we might see the space of explanations (or Explanatory Space \cite{sovrano2020modelling}) as a sort of manifold space where every single overview is not user-centred locally, but globally (as an element of a sequence of information, choosable by a user).
	
	The proposed solution builds over the extraction and structuration of an \acf{ES}, intended (as in \cite{sovrano2020modelling}) as the set of all possible explanations (about an explanandum) reachable by a user, through an explanatory process, starting from an initial explanans, via a pre-defined set of actions. 
	According to the model of \citeauthor{sovrano2020modelling}, we might see the \ac{ES} as a graph of interconnected bits of explanation, and an explanation as nothing more than a path within the \ac{ES}. 
	The aforementioned interconnected bits of explanation can be seen as overviews about different aspects of the explanandum.
	We can see an explanandum aspect as an information cluster. 
	Assuming that the explanandum is a set of documents written in a natural language (e.g. English), the core aspects to explain might be (for example) the different concepts/entities within the corpus, so that to each concept it is possible to associate an overview; e.g. in the sentence \quotes{the customer opened a new bank account} different entities are \quotes{customer}, \quotes{bank}, \quotes{bank account}.
	The choice of an initial explanans is generally dependent on the nature of the explanandum and the objectives associated with the category of users involved in the explanatory process. A good choice of initial explanans could be an overview of the whole explanandum or of the explanatory process. Therefore, in the case of \ac{XAI}, a proper initial explanans might be the static explanation provided by the \ac{XAI} algorithm (e.g. by compiling a template or generating text through a formal language).
	An explanatory process can be defined by the choice of appropriate heuristics for exploring the \ac{ES} and structuring information clusters. 
	Considering that an \ac{ES} is a particular type of graph, and considering that many instances of NP-difficult problems on graphs can be efficiently solved via tree decomposition, the heuristics in question shall provide a policy for at least:
	\begin{inparaenum}[i)]
		\item organising the \ac{ES}'s nodes or aspects,
		\item structuring the information internal to the \ac{ES}'s nodes,
		\item ordering/filtering the \ac{ES}'s edges in a way that would effectively decompose the graph into a tree.
	\end{inparaenum}

\begin{figure}[t]
		\centering
		\includegraphics[width=.9\columnwidth]{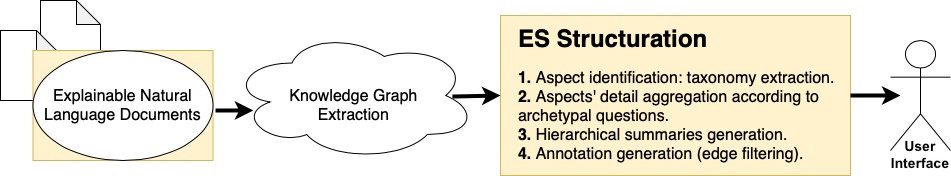}
		\caption{\textbf{The Pipeline}: A simple diagram summarising the pipeline of our user-centric explanatory software.}
		\label{fig:yai_diagram}
	\end{figure}

	The heuristics we adopted are respectively:
	\begin{inparadesc}
    	\item abstraction (for picking the explanandum's aspects, or \ac{ES}'s nodes),
    	\item relevance (for organising the information internal to the \ac{ES}'s nodes)
    	\item and simplicity (for filtering the information internal to \ac{ES}'s nodes and also for selecting the viable \ac{ES}'s edges).
	\end{inparadesc}
	In order to implement the three aforementioned heuristics, for structuring and exploring of the \ac{ES} we need an algorithm that (for example) would, as shown in figure \ref{fig:yai_diagram}:
	\begin{enumerate}
		\item Identify and extract out of the explanandum all the different aspects (concepts/entities) and their related information.
		\item Build a knowledge graph so that concepts/entities are linked together.
		\item Extract a taxonomy from the knowledge graph.
		\item Build one or more information clusters for every aspect, according to the identified archetypal questions.
		\item Present information clusters through a hierarchy of expandable summaries.
		\item Filter the external edges of the \ac{ES}, favouring shorter and simpler paths/explanations, thus reasonably reducing the amount of redundant information for a human.
	\end{enumerate}

	\subsection{Knowledge Graph Extraction} \label{sec:kg_extraction}
	\ac{KG} extraction is the extraction of concepts and their relations, from natural language text, in the form of a graph where concepts are nodes and relations are edges. 
	We are looking for a way to extract \ac{KGs} that somehow preserve the original natural language, preferring them over classical \ac{RDF} graphs. This way we can easily make them inter-operate with deep-learning based \ac{QA} algorithms and existing language models.
	More in detail, as in \cite{sovranolegal}, we perform \ac{KG} extraction by:
	\begin{enumerate}
		\item Analysing the grammatical dependencies of the tokens extracted by Spacy's Dependency Parser, thus identifying the (target) concepts and entities in the form of syntagms.
		\item Using the dependency tree to extract all the tokens connecting two different target concepts in a sentence, thus building a textual template formed by the ordered sequence of the identified tokens and the target concepts replaced with the placeholders \quotes{\{subj\}} and \quotes{\{obj\}} (in accordance with their grammatical dependencies).
		\item Creating a graph of subject-predicate-object triples where the target concepts are the subject and the object and the textual template is the predicate.
	\end{enumerate}
	The resulting triples are a sort of function, where the predicate is the body of the function and the object/subject are its parameters. Obtaining a natural language representation of these template-triples is straightforward by design, by replacing the instances of the parameters in the body. An example of such a template-triple (in the form subject, predicate, object) is:
	\begin{inparadesc}
		\item \quotes{the applicable law},
		\item \quotes{Surprisingly \{subj\} is considered to be clearly more related to \{obj\} rather than to something else.},
		\item \quotes{that Member State}.
	\end{inparadesc}
	Therefore, to increase the interoperability of the extracted \ac{KG} with external resources we performed the following extra steps:
	\begin{inparaenum}[i)]
		\item We automatically assigned a URI and a RDFS label to every node of the graph. The URI is obtained by lemmatising the label.
		\item We automatically added special triples to keep track of the snippets of text (the sources) from which the concepts and the relations are extracted.
		\item We automatically added sub-class relations between composite concepts (syntagms) and the simpler concepts composing the syntagm.
	\end{inparaenum}

	Because of the adopted extraction procedure, the resulting \ac{KG} is not perfect, containing mistakes caused by wrong dependency assignments or similar issues. Despite this, due to the fact that the original natural language is practically preserved thanks to the textual templates, this will not impact significantly on \ac{QA}.
	
	\subsection{Taxonomy Construction: Nodes Clustering} \label{sec:taxonomy_construction}
	In order to efficiently use, query and explore the extracted \ac{KG}, we need to structure it in a proper way.
	We believe that effective abstract querying can be possible by structuring the \ac{KG} as a light ontology, giving it a solid backbone in the form of a taxonomy. In fact, being able to identify the types/classes of a concept would allow to perform queries with a reasonable level of abstraction, making possible to refer to all the sub-types (or to some super-types) of a concept without explicitly mentioning them.
	
	The taxonomy construction phase consists in building one or more taxonomies, via \acf{FCA} \cite{ganter2012formal}.
	In order to build a taxonomy via \ac{FCA} one approach consists in exploiting, as \ac{FCA}'s properties, the hypernyms relations of the concepts in the \ac{KG}. We found that the simplest way to extract such relations is through the alignment of the \ac{KG} to WordNet\footnote{We are aware that WordNet is not omni-comprehensive, but at this stage of the work we are only interested in extracting a reasonable taxonomy.}, through a Word-Sense Disambiguation algorithm.
	The application of \ac{FCA} on the aligned WordNet concepts (and their respective hypernyms) produces as result a forest of taxonomies. Every taxonomy in the forest is a cluster of concepts rooted into very abstract WordNet concepts that we can use as label for the respective taxonomies.
	
	\subsection{Overview Generation via Question Answering: Information Clustering and Summarisation} \label{sec:qa}
	As mentioned in the previous sections, we can generate an overview by clustering and ordering information with respect to its pertinence to a set of archetypal questions.
	The archetypal questions we considered are:
	\begin{inparadesc}
		\item why (standing for causal or justificatory explanations),
		\item what for (teleological),
		\item how (expository),
		\item when (temporal),
		\item where (spatial),
		\item what and who (descriptive).
	\end{inparadesc}
	The essential idea is to generate a concept overview by performing \ac{KG}-based question answering, retrieving the most similar concept's triples for each archetype.
	\ac{KG}-based question answering consists in answering natural language questions about information contained in the \ac{KG}. More in detail, let $Q$ be an archetypal question and $C$ a concept, we perform information clustering by:
	\begin{enumerate}
		\item \textbf{Extracting} all the template-triples related to $C$, including those of $C$'s sub-classes.
		\item \textbf{Selecting}, among the natural language representations of both the retrieved triples and their respective subjects/objects, the snippets of natural language that are sufficiently likely to be an answer to $Q$.
		\item \textbf{Returning} as set of answers the contexts (the source paragraphs) of the selected triples, ordered by pertinence.
	\end{enumerate}
	More in detail, the \textit{selecting} phase is performed by computing the pertinence of an answer as the inner product between the embeddings of the contextualised snippets of text and the embedding of $Q$. The aforementioned embedding is obtained by means of a specialised language model such as the \acf{USE} for \ac{QA} \cite{yang2019multilingual}, while the context is the source paragraph from which a snippet of text is extracted from the original document. If a snippet of text has a similarity above a given threshold, then it is said to be sufficiently likely an answer to $Q$, therefore pertinent.
	
	Considering that an answer could be reasonably associated to more than one archetypal question, we decided to apply an heuristic filtering strategy in the attempt to minimise redundant information, thus following the simplicity heuristic. To do so, we had to attempt a sort of hierarchical organisation of the archetypes, defining some questions as more generic than others, thus prioritising the less generic ones. In fact, in some cases an answer to the question \quotes{What?} could also be a valid answer to \quotes{What for?}. This is because \quotes{What for?} is intuitively more specific than \quotes{What?}. Hence, to reduce redundancy, we can force answers to be exclusive to a single archetypal question, assigning first the answers to the most specific archetypes.
	A descending ordering of specificity, that we found meaningful for the identified archetypal questions, is: 
	\begin{inparadesc}
		\item why,
		\item what for,
		\item how,
		\item who,
		\item where,
		\item when,
		\item and what.
	\end{inparadesc}
	Such ordering seemed to be proper for the purposes of the proof of concept presented in Section \ref{sec:proof_of_concept}, but it is likely that a different ordering is required for different purposes.

	Finally, after the identification of a set of answers for a question $Q$, we can build an expandable summary by recursively concatenating together few answers and by summarising them (thus recursively building a tree of summaries) through one of the state-of-the-art deep learning algorithms for extractive or abstractive summarisation provided by \citeauthor{wolf2019transformers}\cite{wolf2019transformers}.
	At the end of the process we have that an overview is defined by a (sometimes empty) expandable summary for every archetypal question, plus the list of super-classes, sub-classes, sub-types (if any) and eventually few other external resources considered to be of any use (e.g. a short abstract of few words).
	Therefore, we have that the additional taxonomical information is used to guarantee the abstraction policy, while the rest of the information is meant to guarantee both the relevance and the simplicity policies.
	
	\subsection{Overview Annotation: Edge Filtering} \label{sec:overview_annotation}
	Every sentence in the overview is annotated. Annotations consist in linking a concept's embodiment to its corresponding overview (so that clicking on the link would open the overview). 
	The edge filtering algorithm has to decide which syntagms to annotate, in order to avoid annotating every possible concept expressed in a sentence, including redundant or useless ones. More precisely, the edge filtering algorithm would remove:
	\begin{itemize}
		\item Those concepts that can be assumed of scarce relevance for a common user, as those likely to be already known by someone with a basic understanding of English (examples are: day, time, space, November, etc..). These concepts are associable to generic world-knowledge and they can be heuristically identified by analysing the words frequency in the Brown corpus \cite{francis1979brown} or similar corpora.
		\item The concepts with a betweenness centrality equal to $0$. In fact, filtering these concepts would reduce the average length of an explanation (intended as a path over the \ac{ES}) without preventing the user from reaching the information it needs.
	\end{itemize}
	
	\section{Related Work} \label{sec:related_work}
	In literature we found many works on \acf{QA}, and only few of them \cite{zou2014natural,zheng2019interactive,cui2019kbqa} were about (\ac{RDF}) \acf{KGs}.
	As comparison to our work, we point to the many state-of-the-art deep-learning based \ac{QA} algorithms implemented by \citeauthor{wolf2019transformers} \cite{wolf2019transformers}. With these algorithms, using the whole explanandum as input context would require an impractical amount of time (try it\footnote{https://huggingface.co/models?filter=question-answering}) for every posed question, in order to obtain very short (e.g. 2-3 words) answers which quality heavily depends on the selected linguistic model. The practical advantage of our approach is that it is capable of selecting the most relevant text fragments in the context, limiting the search for an answer to very few paragraphs rather than the entire corpus, thus improving efficiency by orders of magnitude. Furthermore our approach can effectively handle very generic questions such as the archetypal ones.
	
	In literature we found many works on explanatory tools, and most of them are focused on one-size-fits-all solutions.
	For example, \citeauthor{cai2019effects} \cite{cai2019effects} propose to use different types of figurative explanations to understand which one is the best, assuming there is always one type of explanation that is the best (or most specific) in average. This attempt to build a sort of hierarchy of explanations is close to what we attempted in Section \ref{sec:qa}, but it clearly lacks any explicit connection to \citeauthor{achinstein1983nature}'s model.
	In \cite{kouki2019personalized} there is a more evident attempt to combine different types of explanations, implicitly considering them useful as answers to different questions a user might pose to itself. Despite this, we see there is a wanted lack of depth in the \ac{ES} generated by \citeauthor{kouki2019personalized}'s tool, thus explanations are shallow and cannot be further expanded.
	Differently from \cite{di2016rank}, we do not re-organise questions on the go, thus answers to the same (archetypal) questions are the same for every user. Despite this, our system is still capable of taking into account the inevitable interest drift pointed by the user during knowledge acquisition. In this sense our approach is a special kind of organisation-based explanation shown to be highly effective by \citeauthor{pu2006trust} in \cite{pu2006trust}.
	
	\section{Proof of Concept Evaluation} \label{sec:proof_of_concept}
	In order to investigate the usefulness of the proposed solution, we designed a user study on a \ac{XAI}-based credit approval system. 
	Being interested in modelling an explanatory process for producing user-centric explanatory software and in quantifying the difference it bears in terms of effectiveness with respect to non-pragmatic approaches, our solution draws from state-of-the-art philosophical theories of explanation.
	What we show is that an abstract philosophical theory of explanations such as \citeauthor{achinstein2010evidence}'s can be beneficially implemented into a concrete software, as a question answering process. We do it by theorising that illocution in explaining involves the act of pertinently answering archetypal questions. We then show how estimations of the pertinence of non-pragmatic answers can result in a globally pragmatic (user-centred) explanatory process.
	If our theory is correct, through the identification of a minimal set of archetypal questions, it is possible to obtain a generator of explanatory overviews generic enough to be able to significantly ease the acquisition of knowledge, regardless of the specific user but depending instead on a fairly broad category of selected users (e.g. people able to read and understand English, hence capable of performing common-sense reasoning), thus resulting in a user-centred explanatory tool that is more effective than its non-pragmatic counterpart on the same explanandum.
	In other terms, our hypothesis is that users, with an explanatory tool based on our model, can understand relevant information more effectively than users without it.
	Despite the specificity of our hypothesis we also want to measure the other aspects of usability (efficiency and satisfaction), in order to have a deeper understanding of the explanatory tool.
	
	In short, we adopt the definition of usability as the combination of \emph{effectiveness}, \emph{efficiency}, and \emph{satisfaction}, as per ISO 9241-210, that defines usability as the \quotes{extent to which a system, product or service can be used by specified users to achieve specified goals with effectiveness, efficiency and satisfaction in a specified context of use} \cite{international2010ergonomics}. Effectiveness (\quotes{accuracy and completeness with which users achieve specified goals}) and efficiency (\quotes{resources used in relation to the results achieved. [\dots] Typical resources include time, human effort, costs and materials.}) can be assessed through objective measures (in our case, pass vs. fail at domain-specific questions and time to complete tasks, respectively). Satisfaction, defined as \quotes{the extent to which the user's physical, cognitive and emotional responses that result from the use of a system, product or service meet the user’s needs and expectations}, is a subjective component and it needs a direct confrontation with the user (in our case through the use of the \ac{SUS} questionnaire \cite{brooke2013sus}).	
	
	%
	In order to verify our hypothesis, we test our pipeline on a \ac{XAI}-powered credit approval system \cite{aix3602019} that is using the \quotes{FICO Explainable Machine Learning Challenge} dataset \cite{holterfico} and probe into it from the perspective of different users. More in detail, we compare a static explanatory tool for \textit{post-hoc} explanations named \acf{CEM} \cite{dhurandhar2018explanations}, with an interactive version built upon our model and that is based on the same informative contents easily reachable by any user of the static explanatory tool (e.g. searching on the web, reading the output of the tool, etc..).
	The aforementioned interactive explanatory tool is specifically designed to be an extension of its static version, so that comparing the usability of those two tools would indirectly allow us to measure the usability of the overviews generator presented throughout the whole paper.
	
	\subsection{The Credit Approval System} \label{sec:experiment_environment}
\begin{figure}[t]
		\centering
		\includegraphics[width=.85\columnwidth]{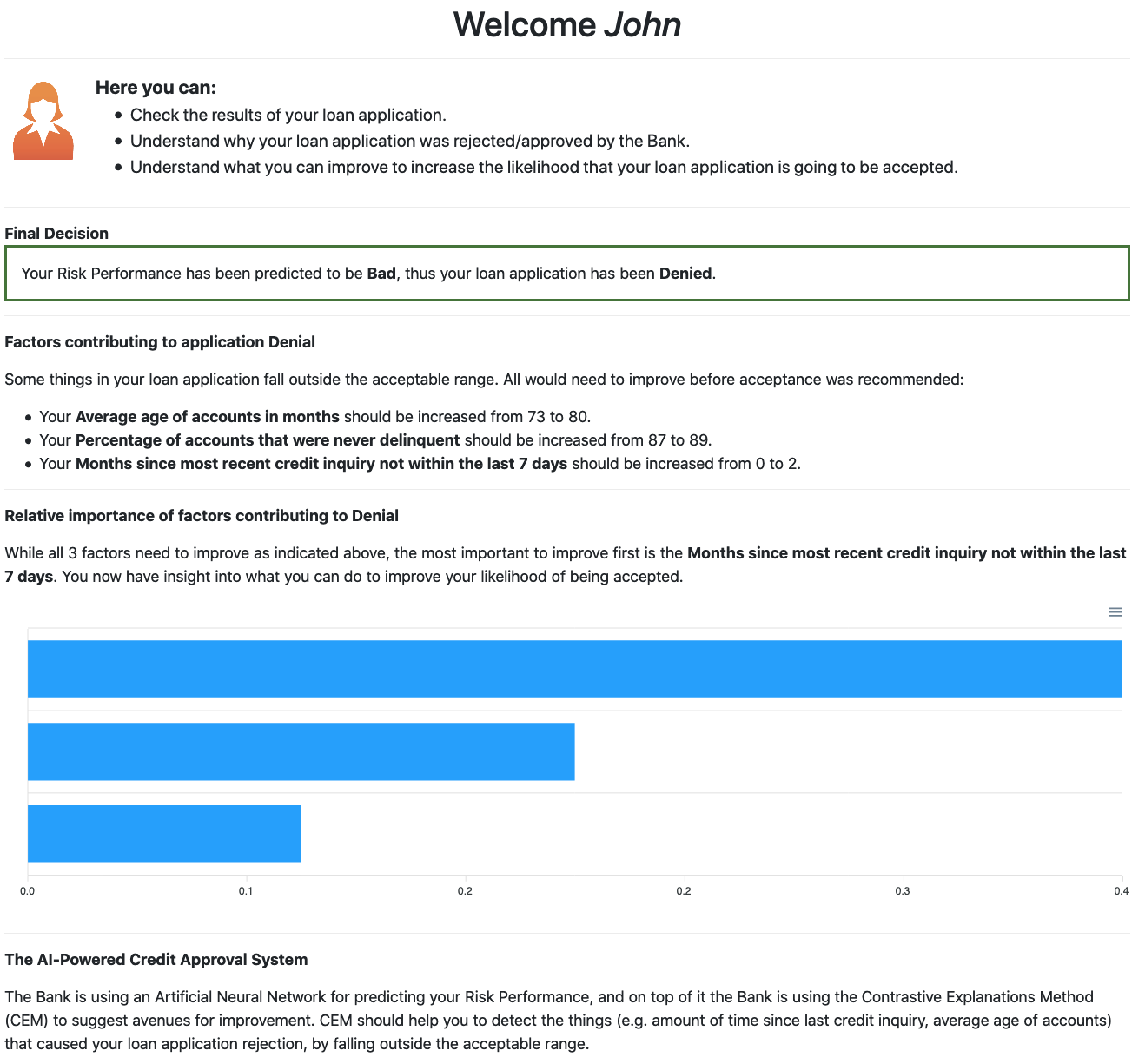}
		\caption{\textbf{Baseline}: A screenshot of the credit approval system for bank customers used as baseline for the proof of concept.}
		\label{fig:baseline_app}
	\end{figure}

	As the context for our evaluation, we took a credit approval system for bank customers, designed by IBM, that is using a static explanatory tool for \textit{post-hoc} explanations named \acf{CEM} \cite{dhurandhar2018explanations}.
	We created a web-page mimicking, with high fidelity, the original IBM's system \cite{aix3602019}, thanks to the available documentation. In Figure \ref{fig:baseline_app} we see an example of static explanation (the initial explanans) produced by the (baseline) credit approval system. 
	A scenario was introduced to the study's participants.
	In this scenario the customer of a bank (John) wants to get an explanation about its loan application, in order to understand why it has been denied by the automated credit approval system.
	The bank uses an artificial neural network, as the automatic process, to decide whether to approve the loan request or not, and it uses the \ac{CEM} algorithm to create a contrastive explanatory information filled into a textual template provided as explanation. This explanation aims at helping customers understand whether they have been treated fairly, providing insights into ways to improve their qualifications so as the likelihood of a future acceptance can be increased.
	
\begin{figure}[t]
		\centering
		\includegraphics[width=.7\columnwidth]{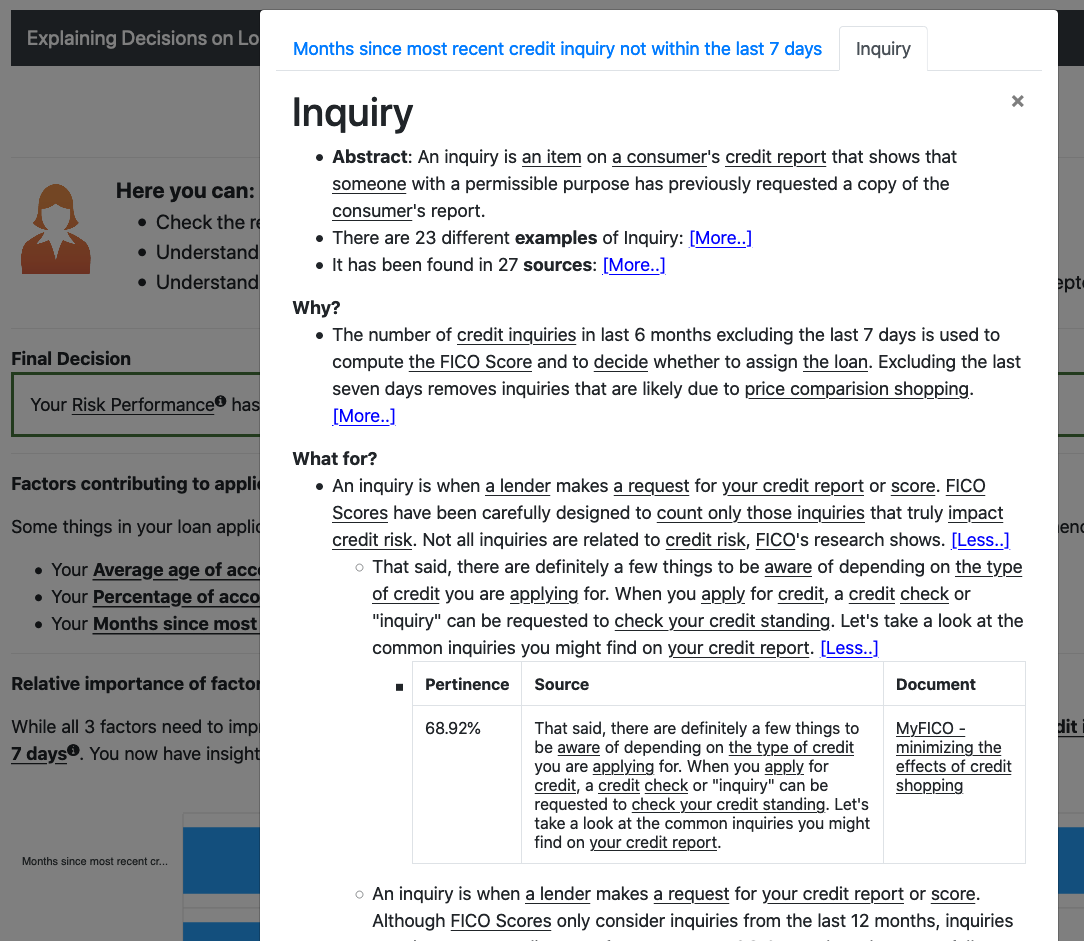}
		\caption{\textbf{Proof of Concept}: Example of overview displaying relevant information about a concept that is directly involved in the initial explanans.}
		\label{fig:alternative_app_overview}
	\end{figure}

	As a comparison with the baseline explanatory tool, we created an interactive version of it that is using our pipeline for generating explanatory overviews\footnote{Source code and additional material is available at \url{https://github.com/Francesco-Sovrano/From-Philosophy-to-Interfaces-an-Explanatory-Method-and-a-Tool-Inspired-by-Achinstein-s-Theory-of-E}, for reproducibility purposes.}.
	This interactive version is obtained by means of a javascript module that automatically annotates the static explanation generated by the baseline tool, making it interactive. 
	As result, the user can click on annotations thus opening an overview modal containing annotations clickable as well, as shown in Figure \ref{fig:alternative_app_overview}.
	The content of the overview modal is obtained by the system by interrogating a python server exposing the necessary APIs to interact with the pipeline described in Section \ref{sec:proposed_solution}.
	The overall extension is designed to be as generic as possible, in other terms it would be possible to use it on any explanatory system providing textual explanations and rich enough documentation (as IBM's), because the aforementioned annotation process is fully automated, as described in Section \ref{sec:proposed_solution}.
	The documentation we used for building the \ac{ES} was taken mainly from \quotes{myfico.com} (50 web-pages) and IBM's website (10 web-pages), plus a few extra web-pages from \quotes{forbes.com}, \quotes{bankrate.com} and Wikipedia, that overall resulted in roughly 57.000 different triplets.
		
	\subsection{User Study} \label{sec:user_study}
	We recruited 103 different participants (57 males, 44 females, 2 unknowns, ages 18-55) on the online platform Prolific \cite{palan2018prolific}. All the participants were recruited among those who:
	\begin{inparaenum}
		\item are resident in UK, US or Ireland;
		\item have a Prolific's acceptance rate greater or equal to 75\%.\footnote{Mainly because they are unlikely to answer poorly/randomly to questions.}
	\end{inparaenum}
	Participants were randomly allocated to test only one of the two versions of the credit approval system: either with (Group 2) or without (Group 1) our extension. 51 participants ended in Group 1 and 52 in Group 2.
	Participants were invited to answer two sets of questions, in English. Participants were told that answering those questions would have taken an average time of 4 minutes. 
	The first set of questions (S1) was domain-specific, so as to measure the \textit{effectiveness} and \textit{efficiency} of the explanations. 
	Three participants were discarded, by answering (more or less) randomly/nonsensically to most of S1.
	At the end of the attention-check process, 97 participants were kept. 
	The questions in S1 are:
	\begin{inparaenum}
		\item What did the Credit Approval System decide for John?
		\item What is an inquiry (in this context)?
		\item What type of inquiries can affect John's score, the hard or the soft ones?
		\item Provide an example of hard inquiry.
		\item How can an account become delinquent?
		\item Which specific process was used by the Bank to automatically decide whether to assign the loan?
		\item What are the known issues of the specific technology used by the Bank (to automatically predict John's risk performance and to suggest avenues for improvement)?
	\end{inparaenum}

	The answers to these questions were scored by a human evaluator as correct (score 1) or not (score 0). For example, given the 1st question, a common correct answer was \quotes{Denial} and a common wrong one was \quotes{The AI-Powered Credit Approval System}.
	In order to ensure that both the explanatory systems (the user-centred one and the baseline) were on an explanandum based on the same informative contents, the participants of Group 1 were explicitly allowed to search the web for correct answers, and many of them did, as shown by the good results on effectiveness of Group 1.
	Furthermore, as discussed also in Section \ref{sec:results}, questions 2, 3, 4, 5 and 7 have been designed so that providing the correct answers would require the exploration of at least 2 or 3 different overviews. 
	
	The second set of questions (S2) is a modified \acf{SUS} questionnaire \cite{brooke2013sus} used to measure the participants' \textit{Satisfaction} with the explanatory systems. The original \ac{SUS} questionnaire has been modified, in order to specifically target the usability of the explanatory systems, by changing every occurrences of the word \quotes{system} with the word \quotes{explanatory system}. Participants were asked to answer S2 by considering only the effort they had for answering S1.
	Of the 97 selected participants: 
	\begin{itemize}
		\item 48 (27 males, 20 females, 1 unknown, ages 18-55) tested the baseline system (V1);
		\item 49 (24 males, 24 females, 1 unknown, ages 18-53) tested the system extended with our pipeline (V2).
	\end{itemize}
	
	\section{Results} \label{sec:results}
	\begin{figure}[t]
		\centering
		\includegraphics[width=.75\columnwidth]{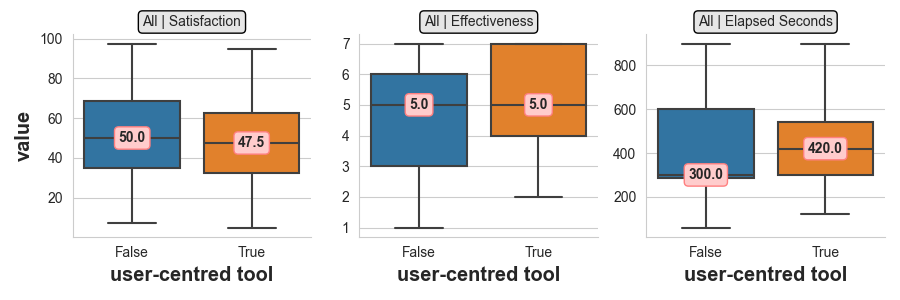}
		\caption{Results are shown in the form of box plots (25th, 50th, 75th percentile, and whiskers covering all data and outliers). The numerical value of medians is shown inside pink boxes. Results for V2 are in orange, while results for V1 are in blue.}
		\label{fig:poc_results_boxplot}
	\end{figure}
	Due to the limited number of samples, we choose to not make assumptions of parametrisation in the data\footnote{Anonymised data is available at \url{https://github.com/Francesco-Sovrano/From-Philosophy-to-Interfaces-an-Explanatory-Method-and-a-Tool-Inspired-by-Achinstein-s-Theory-of-E}, for reproducibility purposes.}. Every sample is grouped according to the adopted version of the credit approval system: V2 is the system using our user-centred explanatory tool; V1 is the baseline. 
	We defined our results measures as:
	\begin{inparadesc}
		\item \textbf{Satisfaction},
		\item \textbf{Effectiveness},
		\item and \textbf{Elapsed Seconds} (that are inversely proportional to Efficiency).
	\end{inparadesc}
	In Figure \ref{fig:poc_results_boxplot} we show the resulting box-plots for every given measure. According to these box-plots, results seem to indicate that V2 is likely to be more effective than V1, but less satisfying.
	We then performed a one-sided Mann-Whitney U-Test (MW) on the other scores, under the alternative hypothesis that the elapsed seconds the and effectiveness of V2 is greater than V1 and the satisfaction of V2 is lower than V1.
	The results show that the global (between-subjects):
	\begin{itemize}
		\item \textbf{Effectiveness score} of V2 is statistically greater than V1. According to MW (U=931.0, p=0.036) there is enough statistical evidence\footnote{Assuming p < 0.05 is enough.}.
		\item \textbf{Satisfaction score} in V2 seems lower than V1, but there is not enough statistical evidence, according to MW  (U=1272.0, p=0.245).
		\item \textbf{Elapsed Seconds} in V2 seems greater than V1, but there is not enough statistical evidence, according to MW  (U=1017.5, p=0.125).
	\end{itemize}
	The obtained results highlight a poor correlation between objective (effectiveness) and subjective (satisfaction) metrics, an interesting phenomenon that has been object of study in many works: \cite{hornbaek2006current,frokjaer2000measuring,nielsen2012user}. 
	We believe that a lower satisfaction is due to the learning process required during the phase of understanding an explanation. In fact, learning can require high cognitive resources. The median amount of time spent in learning could be estimated by computing the difference between the medians of the E
	elapsed seconds of V2 and V1, and that is roughly 120 seconds (definitely something).
	Filtering information and analysing it can be a mentally onerous process. Given the evident intrinsic link between explaining and teaching (and therefore learning) it cannot be excluded that, for a user not strongly interested in achieving the objectives defined for the explanatory process (and therefore a user who is paid to carry out a usability test, like ours), a sufficiently rich explanatory tool is perceived as less satisfactory (because the less time they take to finish the test, the better is for them) than an explanatory tool that follows a one-size-fits-all approach (thus hiding too much complexity, reducing the effectiveness of the system).
	This intuition is supported by a statistically significant difference between the answers to the question number 8 of the \ac{SUS}, where the hypothesis that the information provided by our explanatory tool is more cumbersome than the baseline is confirmed by a MW test (U=822.0, p=0.004). 
	In other words, our tool provides a more effective exploration of the \ac{ES}, but it fails to satisfactorily hide all its complexity, according to the participants. In fact, the archetypal questions used to structure the \ac{ES} are very generic and the user shall choose those most relevant to him. This phase of choosing questions (and therefore explanations) is mentally expensive (even if working with summaries) and it requires having to read snippets of explanation perhaps not useful for the task.
	Said that, the exploratory system we proposed could be suitably designed to facilitate an efficient retrieval of answers to pre-defined questions (such as those of the effectiveness questionnaire) thus minimizing the number of clicks and consequently increasing the overall satisfaction. But to properly evaluate our hypothesis, we decided to design the evaluation questionnaires in such a way that providing the correct answers\footnote{To questions 2,3,4,5, and 7 of the effectiveness questionnaire.} would have required a minimum of \ac{ES} exploration, i.e. the exploration of at least 2 or 3 different overviews. 
	This, obviously, affected the average satisfaction and efficiency but allowed us a better comparison with the baseline and to avoid studying the optimal case (which, by definition, would have given optimal results, but useless for the scientific purposes of the paper).
	
	\section{Conclusions and Future Work} \label{sec:conclusions}
	In this paper, we proposed a new method for explanations in \ac{AI} and, consequently, a tool to test its expressive power within a user interface.
	Being interested in modelling an explanatory process for producing user-centric explanatory software, and in quantifying the difference it bears in terms of effectiveness with respect to non-pragmatic approaches, our solution drawn from state-of-the-art philosophical theories of explanation.
	Among the few philosophical theories of explanation, we identified the one that, we believe, is mostly convertible into a practical model for user-centric explanatory software: \citeauthor{achinstein1983nature}'s.
	But \citeauthor{achinstein1983nature}'s is an abstract illocutionary theory of explanation, therefore we proposed a way to concretely implement illocution as the act of \textit{pertinently answering} archetypal questions (e.g. Why? What for? How? When? etc..), removing from the \quotes{equation} the need for a model of human intention.
	What we showed is that an abstract philosophical theory of explanations can be beneficially implemented into a concrete software, as a question answering process.
	In fact, through the identification of a minimal set of archetypal questions, it is possible to obtain a generator of explanatory overviews generic enough to be able to significantly ease the acquisition of knowledge, regardless of the specific user but depending instead on a fairly broad category of selected users, thus resulting in a user-centred explanatory tool that is more effective than its non-pragmatic counterpart on the same explanandum.
	In other terms, our hypothesis was that users, with an explanatory tool based on our model, can understand relevant information more effectively than users without it.
	In order to test our hypothesis, we had to invent a new pipeline of \ac{AI} algorithms (briefly summarised in figure \ref{fig:yai_diagram}) and run a user study on it. This pipeline is able to organize the information contained in non-structured documents written in natural language (e.g. web pages, pdf, etc..), allowing efficient information clustering, according to a set of archetypal questions, aiming to build a sufficiently rich and effectively explorable \acf{ES} for the automated generation of user-centred explanations. 
	We tested our hypothesis on a \ac{XAI}-powered credit approval system \cite{aix3602019}, comparing a static explanatory tool for \textit{post-hoc} explanations named \acf{CEM} \cite{dhurandhar2018explanations}, with its interactive version based on our model.
	The results of the user study, involving more than 100 participants, showed that our proposed solution produced a statistically relevant improvement on effectiveness (U=931.0, p=0.036) over the baseline, thus giving evidence in favour of our hypothesis.
	
	As a future development, we therefore plan to test our work against other baselines, and to integrate a more classical \ac{QA} system into the system presented here. If the user could ask more specific questions (also obtaining the appropriate answers), then he could save himself part of the effort to search for information, through a more natural mechanism. This intuition is also supported by the findings of \citeauthor{madumal2019grounded} \cite{madumal2019grounded}. In general, however, it is not always possible to evaluate the effectiveness of a more sophisticated \ac{QA} system through the use of a questionnaire, because the first thing a user would do would be asking the questions in the questionnaire.
	
	\bibliographystyle{ACM-Reference-Format}
	\bibliography{biblio}
	
	\appendix
	
%
%
	
	\begin{acronym}
		\acro{EU}{European Union}
		\acro{ADM}{Automated Decision-Making system}
		\acro{ADMs}{Automated Decision-Making systems}
		\acro{AI-HLEG}{High-Level Expert Group on Artificial Intelligence}
		\acro{AI}{Artificial Intelligence}
		\acro{XAI}{eXplainable AI}
		\acro{YAI}{explanatorY AI}
		\acro{HCI}{Human-Computer Interaction}
		\acro{RL}{Reinforcement Learning}
		\acro{EN}{Explanatory Narrative}
		\acro{ENs}{Explanatory Narratives}
		\acro{EP}{Explanatory Process}
		\acro{ES}{Explanatory Space}
		\acro{GDPR}{General Data Protection Regulation}
		\acro{ETTAI}{Explanatory Tool for Trustworthy AI}
		\acro{EI}{Explainable Information}
		\acro{XP}{eXplainable Processes}
		\acro{XD}{eXplainable Datasets}
		\acro{UI}{User Interface}
		\acro{RDF}{Resource Description Framework}
		\acro{AIX360}{AI Explainability 360}
		\acro{CEM}{Contrastive Explanations Method}
		\acro{UCET}{User-Centred Explanatory Tool}
		\acro{SET}{Static Explanatory Tool}
		\acro{KB}{Knowledge Base}
		\acro{TFIDF}{Term Frequency–Inverse Document Frequency}
		\acro{USE}{Universal Sentence Encoder}
		\acro{SUS}{System Usability Scale}
		\acro{EU}{European Union}
		\acro{ADM}{Automated Decision-Making system}
		\acro{ADMs}{Automated Decision-Making systems}
		\acro{AI-HLEG}{High-Level Expert Group on Artificial Intelligence}
		\acro{AI}{Artificial Intelligence}
		\acro{XAI}{eXplainable AI}
		\acro{HCI}{Human-Computer Interaction}
		\acro{RL}{Reinforcement Learning}
		\acro{GDPR}{General Data Protection Regulation}
		\acro{EI}{Explainable Information}
		\acro{XP}{eXplainable Processes}
		\acro{XD}{eXplainable Datasets}
		\acro{UI}{User Interface}
		\acro{RDF}{Resource Description Framework}
		\acro{KG}{Knowledge Graph}
		\acro{KGs}{Knowledge Graphs}
		\acro{TFIDF}{Term Frequency–Inverse Document Frequency}
		\acro{USE}{Universal Sentence Encoder}
		\acro{OKE}{Open Knowledge Extraction}
		\acro{NLP}{Natural Language Processing}
		\acro{FCA}{Formal Concept Analysis}
		\acro{QA}{Question Answering}
		\acro{ODP}{Ontology Design Pattern}
		\acro{ODPs}{Ontology Design Patterns}
		\acro{PIL}{International Private Law}
	\end{acronym}
\end{document}